\documentclass[pdflatex,sn-mathphys-num]{sn-jnl}% Math and Physical Sciences Numbered Reference Style
\usepackage{graphicx}%
\usepackage{multirow}%
\usepackage{amsmath,amssymb,amsfonts}%
\usepackage{amsthm}%
\usepackage{mathrsfs}%
\usepackage[title]{appendix}%
\usepackage{xcolor}%
\usepackage{textcomp}%
\usepackage{manyfoot}%
\usepackage{booktabs}%
\usepackage{algorithm}%
\usepackage{algorithmicx}%
\usepackage{algpseudocode}%
\usepackage{listings}%
\theoremstyle{thmstyleone}%
\theoremstyle{thmstyletwo}%

\theoremstyle{thmstylethree}%

\begin{document}

\title[Article Title]{
Calibrating Agent-Based Financial Markets Simulators with Pretrainable Automatic Posterior Transformation‑Based Surrogates}


\author[1]{\fnm{Boquan} \sur{Jiang}}\email{12332460@mail.sustech.edu.cn}

\author[1]{\fnm{Zhenhua} \sur{Yang}}\email{yangzh2022@mail.sustech.edu.cn}

\author[2]{\fnm{Chenkai} \sur{Wang}}\email{wangck2022@mail.sustech.edu.cn}

\author[1]{\fnm{Muyao} \sur{Zhong}}\email{12149032@mail.sustech.edu.cn}

\author[1]{\fnm{Heping} \sur{Fang}}\email{12331106@mail.sustech.edu.cn}

\author*[1,2]{\fnm{Peng} \sur{Yang}}\email{yangp@sustech.edu.cn}

\affil[1]{\orgdiv{Guangdong Provincial Key Laboratory of Brain-Inspired Intelligent Computation, Department of Computer Science and Engineering},
          \orgname{Southern University of Science and Technology},
          \orgaddress{\city{Shenzhen},
                      \state{Guangdong},
                      \postcode{518055},
                      \country{China}}}

\affil[2]{\orgdiv{Department of Statistics and Data Science},
          \orgname{Southern University of Science and Technology},
          \orgaddress{\city{Shenzhen},
                      \state{Guangdong},
                      \postcode{518055},
                      \country{China}}}


\abstract{Calibrating Agent-Based Models (ABMs) is an important optimization problem for simulating the complex social systems, where the goal is to identify the optimal parameter of a given ABM by minimizing the discrepancy between the simulated data and the real-world observations. Unfortunately, it suffers from the extensive computational costs of iterative evaluations, which involves the expensive simulation with the candidate parameter. While Surrogate-Assisted Evolutionary Algorithms (SAEAs) have been widely adopted to alleviate the computational burden, existing methods face two key limitations: 1) surrogating the original evaluation function is hard due the nonlinear yet multi-modal nature of the ABMs, and 2) the commonly used surrogates cannot share the optimization experience among multiple calibration tasks, making the batched calibration less effective.
To address these issues, this work proposes Automatic posterior transformation with Negatively Correlated Search and Adaptive Trust-Region (ANTR). ANTR first replaces the traditional surrogates with a pretrainable neural density estimator that directly models the posterior distribution of the parameters given observed data, thereby aligning the optimization objective with parameter-space accuracy. Furthermore, we incorporate a diversity-preserving search strategy to prevent premature convergence and an adaptive trust-region method to efficiently allocate computational resources. We take two representative ABM-based financial market simulators as the test bench as due to the high non-linearity.
Experiments demonstrate that the proposed ANTR significantly outperforms conventional metaheuristics and state-of-the-art SAEAs in both calibration accuracy and computational efficiency, particularly in batch calibration scenarios across multiple market conditions.}




\maketitle

\section{Introduction}\label{sec1}

Agent-Based Modelling (ABM) is a mainstream simulation technique that constructs a population of independent, heterogeneous, parameterized agents and simulates their interactions within a given environment while adhering to specific rules~\cite{farmer2009economy}. These local interactions collectively drive the evolution of the ABM. ABM has been widely applied to model a variety of real-world complex social systems, including consumer behavior~\cite{zhang2007agent}, infectious disease dynamics~\cite{tracy2018agent}, and financial markets~\cite{dosi2013income}.

In the context of financial markets, ABM aggregates price dynamics emerging from the interactions of heterogeneous agents with diverse objectives, information, and trading behaviors. An ABM implements a limit‑order‑book (LOB) market in which heterogeneous agents (e.g., liquidity providers and takers) interact via a continuous double auction (CDA). Agents parameterized by $\boldsymbol{\theta}$ (e.g., order arrival intensities, decision latencies, and order‑size distributions) submit limit orders to a price–time–priority matching engine; unmatched orders remain on the book and constitute the observable market state. This generates a time series of LOB states from which quotes, spreads, and the mid‑price are computed. Running $M(\boldsymbol{\theta})$ for $T_{\mathrm{obs}}$ steps yields a simulated time series $\boldsymbol{x}^{\boldsymbol{\theta}}_{0:T_{\mathrm{obs}}}$ to be compared with observed data $\boldsymbol{x}_{\mathrm{obs}}$ (generated by certain ground-truth parameter $\boldsymbol{\theta}_{\mathrm{true}}$).

Calibrating agent-based models for financial markets refers to the process of optimizing model parameters so that the time series generated by a financial simulator can closely match real-world market observations, e.g., prices over time ~\cite{mcculloch2022calibrating,zhao2025learning,zhong2025representation,wang2025alleviating}. 
This task can be formulated as a black-box optimization problem:
\begin{equation}
    \boldsymbol{\theta}^\ast 
    = \arg\min_{\boldsymbol{\theta} \in \boldsymbol{\Theta}} 
      D\big(M(\boldsymbol{\theta}),\, \boldsymbol{x}_{\mathrm{obs}}\big),
\end{equation}
where $\boldsymbol{\theta}$ denotes the vector of model parameters, $\boldsymbol{\Theta}$ is the feasible parameter space, $M(\boldsymbol{\theta}) = \boldsymbol{x}^{\boldsymbol{\theta}}_{0:T_{\mathrm{obs}}}$ represents the simulated time series of length $T_{\mathrm{obs}}$ generated under parameters $\boldsymbol{\theta}$, $\boldsymbol{x}_{\mathrm{obs}}$ denotes the observed market price data, and $D(\cdot,\cdot)$ is a discrepancy measure between simulated and observed data. A well-calibrated model enables policymakers, financial practitioners, and researchers to conduct reliable analyses for tasks such as policy evaluation~\cite{zhao2025quantfactor}, market intervention~\cite{farmer2009economy}, and risk assessment~\cite{vytelingum2025agent}. Accordingly, the credibility of such analyses largely depends on whether the model parameters have been properly calibrated~\cite{steinbacher2021advances}.

Given that the multi-agent-based simulator $M$ is typically implemented as a complex computer program, the objective in Eq.~(1) is non-differentiable. 
As a result, metaheuristic algorithms~\cite{abdel2018metaheuristic}, such as Particle Swarm Optimization (PSO)~\cite{kennedy1995particle} and Evolutionary Algorithms (EAs)~\cite{stonedahl2010evolutionary}, have been widely adopted for financial ABM calibration~\cite{wang2021calibration}. 
These population-based approaches are well suited for non-convex, non-differentiable, and noisy optimization landscapes, making them attractive for calibrating financial ABMs\cite{lamperti2018agent,platt2020comparison}. However, the practical applicability of these methods is often constrained by computational limitations. Since each function evaluation involves a full ABM simulation run, and robust calibration typically requires thousands of such evaluations, the optimization process becomes prohibitively expensive. For example, the calibration of a five-minute segment of per-second financial data may require thousands of simulator calls, with each call taking approximately 15 seconds, resulting in a total calibration time exceeding 240 minutes~\cite{li2024simlob}. This 240/5=48 times lag between the calibration time and the data generation time in real markets makes the computational costs unacceptable. 

In recent years, Surrogate-Assisted Evolutionary Algorithms (SAEAs) have become a dominant approach to reduce computational cost by introducing computationally cheap surrogate models to replace expensive simulator-based evaluations ~\cite{he2023review}. A typical SAEA optimization framework consists of three core components: first, a small set of candidate solutions (i.e., parameter combinations) are evaluated with the real objective function, and the results are used to train a surrogate model to approximate it; second, EA operates on the surrogate model to search for promising solutions, and the best candidates are selected to guide the next round of sampling; finally, through an iterative cycle of surrogate model updates and sample evaluations, the algorithm can achieve high-quality optimization performance under a controlled evaluation time budget~\cite{liang2024reviewSAEA}.

In practice, SAEAs commonly employ Gaussian Processes (GPs)~\cite{santner2003design}, Radial Basis Functions (RBFs)~\cite{knysh2016blackbox}, or Neural Networks (NNs)~\cite{angione2022using} to construct accurate approximations of the underlying objective landscape. Among these, GPs are particularly popular due to their ability to provide accurate predictions from limited data while offering both mean and variance estimates~\cite{he2023review}.

Despite the advantages of SAEAs, a fundamental limitation persists in the surrogate evaluation loop of the framework. Essentially, most existing approaches focus on approximating a data-level discrepancy $D\!\left(M(\boldsymbol{\theta}^\ast),\, M(\boldsymbol{\theta}_{\mathrm{true}})\right)$, i.e., the fitness evaluation function of the optimization problem. However, due to the highly nonlinear nature of ABMs, a small data-space discrepancy $D\!\left(M(\boldsymbol{\theta}^\ast),\, M(\boldsymbol{\theta}_{\mathrm{true}})\right)$ does not necessarily imply proximity in the parameter space, i.e., $\lVert \boldsymbol{\theta}_{\mathrm{true}} - \boldsymbol{\theta}^\ast \rVert$ may still be large. As a result, both metaheuristic algorithms and SAEAs may converge to parameter configurations that yield realistic-looking data but remain far from the ground-truth parameters~\cite{yang2025towards}. In financial ABM calibration, however, recovering parameters close to their true values is crucial, as they encode key behavioral and structural aspects of the market that govern responses to shocks or policy changes. Significant deviations may still yield realistic-looking data but can lead to misleading results in counterfactual or policy analyses.

Another challenge lies in the fact that the simulators often need to be applied in multiple scenarios and thus the calibration tasks commonly emerge in batches, e.g., calibration financial ABMs across multiple datasets, trading days, or market conditions. In such scenarios, enjoying the calibration experience from existing calibrated scenarios would provide better initialization, reduce the number of expensive simulations required for convergence, and improve stability for the remaining tasks. However, existing surrogates hardly share such experience across tasks~\cite{clarke2005analysis}.

To address the above issues, we propose a pretrained surrogate model that directly models the parameter space via a posterior distribution. Specifically, the model explicitly estimates the posterior distribution
$
p(\boldsymbol{\theta} \mid \boldsymbol{x}_{\mathrm{obs}}),
$
which quantifies the plausibility of a candidate parameter vector $\boldsymbol{\theta}$ being the ground-truth parameter given the observed data $\boldsymbol{x}_{\mathrm{obs}}$.
That is, the higher the posterior probability density, the closer the parameter $\boldsymbol{\theta}$ is to the ground-truth parameter. Conceptually, the posterior distribution learns the mappings of the parameters and the observed data given the ABM-based simulator. And thus can be trained with prepared pairs of supervised signals $<\boldsymbol{\theta}_{\mathrm{true}},\boldsymbol{x}_{\mathrm{obs}}>$. As long as the batch calibration tasks are concentrated on the same ABM, the posterior can be pretrained and shared across multiple data batches. In other words, the surrogate could be trained for a class of calibration tasks regarding the same $M$, rather than for a single instance as existing surrogates do. 

We observe that the Automatic Posterior Transformation (APT)~\cite{greenberg2019automatic}, the state-of-the-art sequential neural posterior estimation technique in the field of Simulation-Based Inference (SBI)~\cite{cranmer2020frontier}, can directly learn the posterior distribution $p(\boldsymbol{\theta}|\boldsymbol{x})$ and support pretraining independent of specific observations. These properties address the data--parameter mismatch in SAEAs and enable efficient reuse across calibration tasks for the same ABM. However, when the traditional Gaussian Process surrogate is simply replaced with APT, the Evolutionary Algorithm (EA) used for sampling tends to focus on expensive simulations in regions where the posterior estimates are high. If the true parameters do not lie within these regions, such bias may compound over iterations, ultimately leading to incorrect optimization results. To maintain effective exploration even when the posterior mass is concentrated in a limited region, we incorporate Negatively Correlated Search (NCS)~\cite{tang2016negatively} for multi-modal exploration. In addition, we employ an adaptive trust-region mechanism that allocates expensive simulations to regions with high posterior probability.

The main contribution of this work is the proposal of a surrogate-assisted optimization framework for the calibration of financial ABMs. This framework integrates the following key components into a unified design:

\textbf{Amortized posterior surrogate:} to address the mismatch between data-space surrogates and parameter-space accuracy, and to overcome the limitations of GP-based pretraining in highly nonlinear financial ABM calibration, we replace the traditional GP surrogate with APT, a pretrainable surrogate model that directly learns the posterior distribution over parameters.

\textbf{Multi-modal
exploration via NCS and adaptive trust-region:} to prevent the search process from collapsing too narrowly around the parameter regions favored by the posterior surrogate, we employ diversity-preserving exploration through NCS combined with an adaptive trust-region mechanism. This design enables broader coverage of the parameter space while still concentrating expensive simulations in promising regions, leading to stable performance improvements over conventional metaheuristics and state-of-the-art in SAEAs.

Our experimental results demonstrate that the proposed Amortized Neural Posterior Estimation with Negatively Correlated Search and Adaptive Trust-Region (ANTR) algorithm significantly outperforms the baseline methods across multiple simulation environments. Under constrained evaluation budgets, ANTR achieves an average reduction of approximately 50\% in parameter recovery error. Furthermore, when matching or surpassing the final performance of the baselines, ANTR requires only 36\%--53\% of their evaluation budget on average, highlighting its superior sample efficiency.

The remainder of this paper is organized as follows. Section~2 introduces the related work. Section~3 presents the framework and detailed components of the proposed ANTR algorithm. Section~4 details the experimental setup, reports the results, and discusses the findings. Finally, Section~5 concludes the paper and outlines directions for future work.

\section{Related Work}\label{sec2}

Traditional EAs typically assume that the objective and constraint evaluations are computationally inexpensive. However, this assumption often breaks down in real-world applications, where fitness evaluations rely on expensive experiments or computationally intensive simulations. Typical examples include airfoil shape optimization based on high-fidelity simulation models~\cite{chen2022inverse}, axial-flow compressor design using computational fluid dynamics (CFD) software~\cite{cheng2019surface}, and parameter calibration of financial agent-based models (ABMs)~\cite{maxe}. In such problems, evaluations are not only expensive but also sometimes unavailable during the optimization process. To address these challenges, data-driven evolutionary algorithms (DDEAs)—also known as surrogate-assisted evolutionary algorithms (SAEAs)—have been proposed~\cite{DDEAoverview1}. These methods construct surrogate models from historical data to partially or fully replace real evaluations, thereby maintaining effective search performance while obtaining satisfactory solutions under a limited budget of true fitness evaluations.

Based on whether additional real evaluations can be obtained during the optimization process, DDEAs can generally be classified into online and offline categories~\cite{huang2021comparative}. Online DDEAs continuously acquire new samples to improve the accuracy of surrogate models while balancing convergence and population diversity throughout the evolutionary process. In contrast, offline DDEAs operate without access to additional evaluations and thus focus on improving the reliability and robustness of surrogate-based candidate predictions.

Among offline DDEAs, representative approaches include DDEA-SE, which pretrains a large surrogate pool and adaptively selects a small yet diverse subset to achieve more accurate local approximations~\cite{wang2018offline}, and MS-DDEO, which introduces a high-level model selection framework that adaptively identifies the most suitable models, including potential ensemble models, to enhance generalization capability and problem adaptability~\cite{zhen2022offline}.

Regardless of whether they follow the online or offline paradigm, most existing methods adopt regression-based surrogates such as Gaussian Process (GP)~\cite{santner2003design}, Radial Basis Function (RBF)~\cite{knysh2016blackbox}, and Neural Network (NN)~\cite{angione2022using} models. In particular, these approaches typically construct a surrogate $\hat{D}(\boldsymbol{\theta})$ to approximate the discrepancy function $D(M(\boldsymbol{\theta}),\, \boldsymbol{x}_{\mathrm{obs}})$, which quantifies the deviation between simulated and observed data and is used to guide the optimization process. Specifically, in the context of surrogate-assisted calibration for agent-based models (ABMs), Lamperti et al.~\cite{lamperti2018agent} employed gradient boosting trees (XGBoost) as a surrogate model to calibrate the Brock–Hommes asset pricing model. Beyond this example, other studies have also applied SAEAs to parameter calibration in ABM tasks, either as part of upstream preprocessing or downstream optimization, thereby achieving more efficient and reliable calibration results~\cite{li2024simlob,bai2022efficient,yang2025towards}. However, such data-discrepancy-based modeling approaches reformulate the objective function of an ABM according to the specific observed data. As a result, whenever the observation changes, new samples must be collected and the surrogate needs to be reconstructed. This dependency on observation-specific objectives hinders the effective reuse of historical samples and limits the ability to leverage pretraining for stably characterizing complex ABMs, thereby reducing the efficiency of batch calibration across datasets or varying conditions for the same model.

In view of these limitations, several studies have instead constructed a surrogate $\hat{M}(\boldsymbol{\theta})$ to approximate the original simulator $M(\boldsymbol{\theta})$, thereby directly replacing its simulation output generation process. For example, Angione et al.~\cite{angione2022using} employed a global machine learning surrogate model to approximate the outputs of a social care ABM. Similarly, Robertson et al.~\cite{robertson2025bayesian} proposed a random forest surrogate that generates fixed-dimensional summary statistics and embeds them into a Bayesian inference pipeline to enable efficient posterior sampling. Although these methods directly model the original simulator $M(\boldsymbol{\theta})$, they still rely mainly on comparing discrepancies either in simulated data or in features extracted therefrom. Such data-level comparisons fail to capture the intrinsic relationship between the estimated and true parameters.

Although the aforementioned DDEAs provide limited support for batch calibration of ABMs, the strategies developed within the online paradigm remain valuable for understanding how to perform efficient search under a limited evaluation budget. Among these methods, Efficient Global Optimization (EGO)~\cite{jones1998efficient} systematically combines Gaussian Process (GP) surrogates with acquisition functions for the global optimization of expensive black-box problems. Similarly, inspired by committee-based active learning, a surrogate-assisted particle swarm optimization algorithm was proposed to intelligently switch between global and local surrogates so as to balance exploration and exploitation~\cite{wang2017committee}. Furthermore, Trust Region Bayesian Optimization (TuRBO)~\cite{eriksson2019scalable} employs multiple adaptive trust regions to preserve the sample efficiency of GPs while significantly improving scalability in high-dimensional and multimodal landscapes. Moreover, Bai et al.~\cite{bai2022efficient} further demonstrated the effectiveness and efficiency of TuRBO for the calibration of multi-agent ABMs.

\section{The Proposed ANTR Framework}\label{sec3}

\subsection{The Proposed Framework}\label{subsec3.1}

\begin{figure}[htbp]
    \centering    
    \includegraphics[width=1\textwidth]{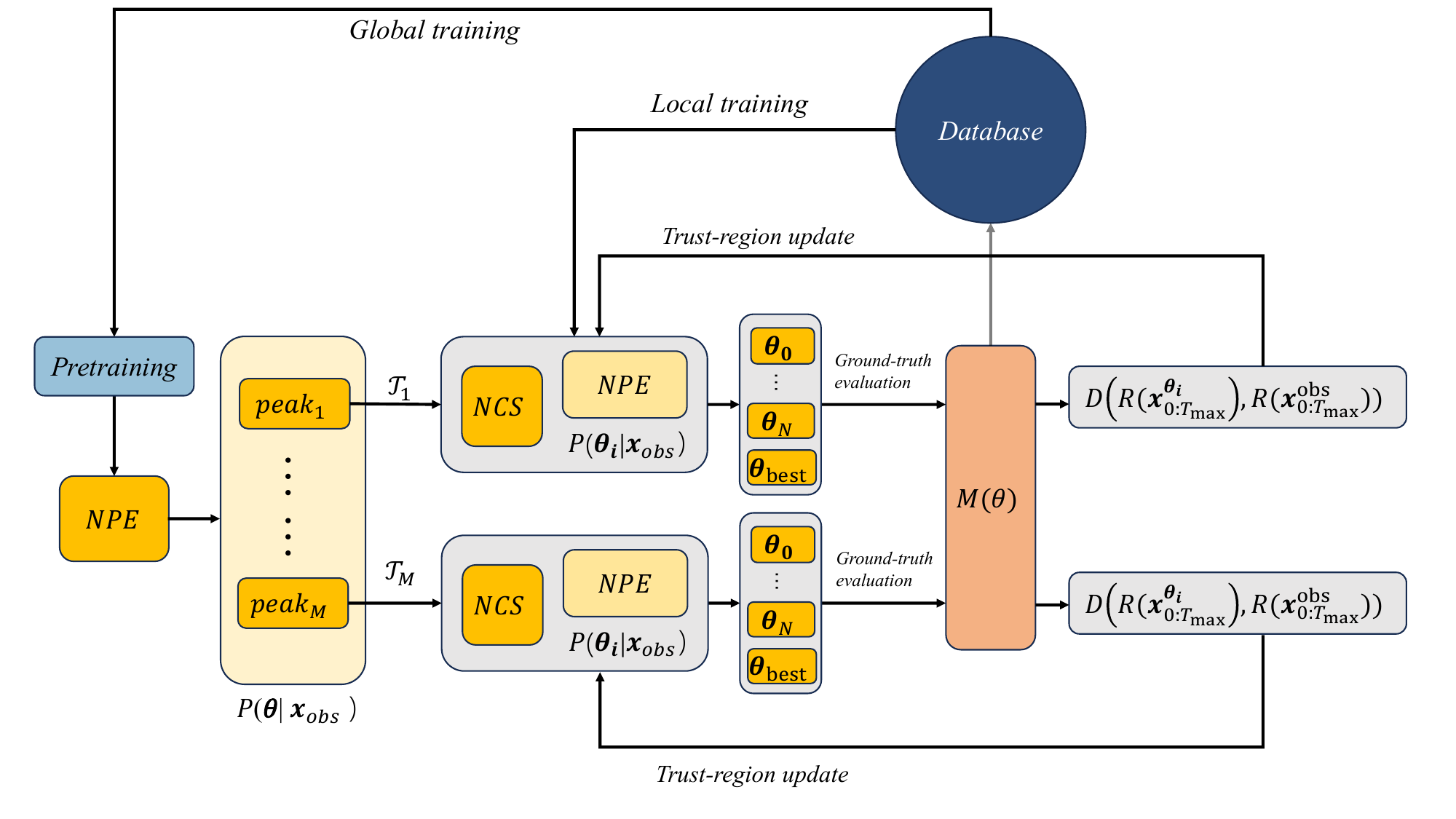}
    \caption{Overall framework of the proposed ANTR framework}
    \label{fig:oursalgorithm}
\end{figure}

To address the high computational costs and strong nonlinearity inherent in financial ABM calibration, we propose the Amortized Neural Posterior Estimation with Negatively Correlated Search and Adaptive Trust-Region (ANTR) algorithm. The pseudocode of the algorithm is presented in Algorithm~\ref{ours}, and the overall framework is illustrated in Fig.~\ref{fig:oursalgorithm}. The method adopts a hierarchical global-to-local optimization paradigm, seamlessly integrating surrogate-assisted evolutionary search with a multi-region trust region mechanism to form a closed-loop, self-adaptive calibration framework.

ANTR consists of three core modules: 
(i) \textbf{Amortized Posterior Density Estimator model}, 
(ii) \textbf{Candidate Generation model}, and 
(iii) \textbf{Trust-Region Adaptation model}. Specifically, the initialization step employs the amortized posterior density estimator to estimate the posterior distribution 
$q_\phi(\boldsymbol{\theta} \mid \boldsymbol{x}_{\mathrm{obs}})$ and identify high-density regions for initializing $M$ independent trust regions 
$\{\mathcal{T}_j\}_{j=1}^M$. For each trust region, the posterior density estimator is trained on the relevant data subsets to obtain region-specific local surrogate model. Based on these surrogates, the candidate generation module applies region-specific search strategies to produce diverse candidate parameters. Finally, the dynamic trust-region adaptation module balances exploration and exploitation via success-history-based adaptive mechanisms while managing the lifecycle of trust regions, including expansion, contraction, creation, merging, and removal.

The detailed design and implementation of each module will be elaborated in the subsequent subsections.

\begin{algorithm}[H]
\caption{ANTR Calibration Algorithm}\label{ours}
\begin{algorithmic}[1]
\Require Global SBI model $\mathcal{M}_{\text{global}}$, simulation database $\mathcal{D}$, target observation $\boldsymbol{x}_{\mathrm{obs}}$, fitness function $\Phi(\boldsymbol{\theta}, \boldsymbol{x}_{\mathrm{obs}})$
\Ensure Optimal local parameters $\{\boldsymbol{\theta}_j^*\}_{j=1}^M$
\State \textbf{Initialization:}
\State Estimate initial posterior $q_\phi(\boldsymbol{\theta} | \boldsymbol{x}_{\mathrm{obs}})$ via $\mathcal{M}_{\text{global}}$
\State Initialize $M$ trust regions $\{\mathcal{T}_j\}_{j=1}^M$ based on high-density regions of $q_\phi(\boldsymbol{\theta} | \boldsymbol{x}_{\mathrm{obs}})$
\State Initialize best solutions $\boldsymbol{\theta}_j^*$
\State Set counters: $c_{s,j} \gets 0, c_{f,j} \gets 0$ for each region $j \in \{1, \dots, M\}$
\For{$t = 1 \dots T$}

    \vspace{0.5em}
    \State \textbf{Local Surrogate Construction:}
    \For{$j = 1 \dots M$}
        \State $\mathcal{D}_j \gets \{(\boldsymbol{\theta}, \boldsymbol{x}) \in \mathcal{D} \mid \boldsymbol{\theta} \in \mathcal{T}_j\}$ \Comment{Retrieve samples in local TR}
        \State Train local surrogate $\mathcal{M}_j$ using $\mathcal{D}_j$
        \State Update local objective $\mathcal{L}_j(\cdot) \gets q_{\phi,j}(\boldsymbol{\theta} | \boldsymbol{x}_{\mathrm{obs}})$ via $\mathcal{M}_j$
        
        \vspace{0.5em}
        \State \textbf{Candidate generation:}
        \State Generate $N$ candidates $\mathbf{\Theta}_{cand}$ via NCS guided by $\mathcal{L}_j(\cdot)$
        \State Evaluate candidates: $\mathcal{X}_{cand} \gets \text{Simulator}(\mathbf{\Theta}_{cand})$
        \State Update database: $\mathcal{D} \gets \mathcal{D} \cup \{(\mathbf{\Theta}_{cand}, \mathcal{X}_{cand})\}$
        \State Find best candidate: $\boldsymbol{\theta}_{best, t} = \arg\max_{\boldsymbol{\theta} \in \mathbf{\Theta}_{cand}} \Phi(\boldsymbol{\theta}, \boldsymbol{x}_{\mathrm{obs}})$
        
        \vspace{0.5em}
        \State \textbf{Trust-Region Adaptation:}
        \If{$\Phi(\boldsymbol{\theta}_{best, t}, \boldsymbol{x}_{\mathrm{obs}}) > \Phi(\boldsymbol{\theta}_j^*, \boldsymbol{x}_{\mathrm{obs}})$} 
            \State $\boldsymbol{\theta}_j^* \gets \boldsymbol{\theta}_{best, t}$ \Comment{Update archival best}
            \State $c_{s,j} \gets c_{s,j} + 1, \quad c_{f,j} \gets 0$
        \Else
            \State $c_{f,j} \gets c_{f,j} + 1, \quad c_{s,j} \gets 0$
        \EndIf
        \If{$c_{s,j} \ge 3$} Expand $\mathcal{T}_j$ and reset $c_{s,j}$ \EndIf
        \If{$c_{f,j} \ge \text{fail}_{\text{tol}}$} Shrink $\mathcal{T}_j$ and reset $c_{f,j}$ \EndIf
    \EndFor \Comment{End of regions loop}
\EndFor \Comment{End of iterations loop}
\State \Return $\{\boldsymbol{\theta}_j^*\}_{j=1}^M$
\end{algorithmic}
\end{algorithm}

\subsection{Amortized Posterior Density Estimator}\label{subsec3.2}

The core functionality of this module is to infer the conditional distribution of the model parameters $\boldsymbol{\theta}$ given the observed data $\boldsymbol{x}_{\mathrm{obs}}$, i.e.,
\begin{equation}
q_\phi(\boldsymbol{\theta} \mid \boldsymbol{x}_{\mathrm{obs}}) \approx p(\boldsymbol{\theta} \mid \boldsymbol{x}_{\mathrm{obs}}),
\end{equation}
where $\boldsymbol{x}_{\mathrm{obs}}$ denotes the observed time series.
In this study, we adopt Automatic Posterior Transformation (APT)~\cite{greenberg2019automatic} as the core amortized posterior density estimator.
APT belongs to the family of Neural Posterior Estimation (NPE) methods within Simulation-Based Inference (SBI) and can be viewed as an extension of conventional NPE approaches.

APT supports the integration of customized embedding networks for feature extraction and can be combined with flexible neural density estimators, such as Mixture Density Networks (MDN) or Normalizing Flows.
These components can be trained jointly in an end-to-end manner, resulting in enhanced flexibility and generalization capability.
As a consequence, APT is particularly well suited to the challenges addressed in this work, including high evaluation costs and the handling of variable-length time series.

To address the issue of variable sequence lengths in financial market calibration tasks, we adopt the following strategy. First, all simulated output sequences are zero-padded to the maximum length $T_{\text{max}}$. Since ABM-generated data typically does not contain zeros in its numerical distribution, the padding can be regarded as a non-informative symbol and thus does not interfere with feature extraction. Second, we employ a convolutional neural network (CNN) to encode the padded sequences into a fixed-dimensional representation $\boldsymbol{z}$. Compared with RNNs or Transformers, CNNs provide faster computation while already delivering satisfactory performance, making it unnecessary to adopt more complex embedding architectures for feature extraction. Finally, the embedding $\boldsymbol{z}$ is fed as a conditional input to the neural density estimation module in the APT framework, allowing the posterior distribution $p(\boldsymbol{\theta} \mid \boldsymbol{z})$ to be learned in an end-to-end manner.

\subsection{Candidate generation}\label{subsubsec3.3}

The primary objective of this module is to exploit the local surrogate models introduced above to efficiently generate candidate solutions for subsequent high-fidelity evaluations. Within this module, Negatively Correlated Search (NCS) is employed as the search mechanism to balance solution quality and diversity. NCS is a search strategy originally proposed by Tang et al. in the context of evolutionary algorithms~\cite{tang2016negatively}. Its core idea is to explicitly encourage diversity among multiple parallel randomized local searches (RLS), such that different search trajectories complement each other and collectively enhance the exploration of the solution space.  

 NCS maintains $N$ parallel RLS processes, each of which can be regarded as a probability distribution $p_i(\boldsymbol{\theta})$. In each iteration, new candidates $\boldsymbol{\theta}_i'$ are sampled from these distributions and evaluated based on both their surrogate-predicted objective value $f(\boldsymbol{\theta}_i')$ and their behavioral difference from other searchers. The latter is quantified using the Bhattacharyya distance:
\begin{equation}
	D_B(p_i, p_j) = \frac{1}{8} (\boldsymbol{\theta}_i - \boldsymbol{\theta}_j)^T \boldsymbol{\Sigma}^{-1} (\boldsymbol{\theta}_i - \boldsymbol{\theta}_j) + \frac{1}{2} \ln \left( \frac{\det \boldsymbol{\Sigma}}{\sqrt{\det \boldsymbol{\Sigma}_i \det \boldsymbol{\Sigma}_j}} \right),
\end{equation}
where $\boldsymbol{\Sigma} = \tfrac{1}{2}(\boldsymbol{\Sigma}_i + \boldsymbol{\Sigma}_j)$. This closed-form expression holds when Gaussian mutation is adopted as the search operator. 

To balance solution quality (objective value) and behavioral diversity, NCS employs the following replacement rule. In each iteration, for every RLS, if
\begin{equation}
	\frac{f(\boldsymbol{\theta}_i)}{\mathrm{Corr}(p_i')} < \lambda,
\end{equation}
the candidate $\boldsymbol{\theta}_i'$ replaces the current solution $\boldsymbol{\theta}_i$, where $\mathrm{Corr}(p_i') = \min_{j \neq i} D_B(p_i', p_j)$ denotes the minimum distance between the new search behavior and all others, and $\lambda$ is a hyperparameter controlling the trade-off between exploration and exploitation.  

Compared with traditional diversity-preserving mechanisms (e.g., sharing functions or crowding distance), NCS promotes negatively correlated search behaviors rather than merely maintaining solution-level distances. Even when identical operators are applied, NCS encourages different RLS processes to explore distinct regions of the search space, thereby sustaining comprehensive exploration and demonstrating strong performance in non-convex optimization problems.  

\begin{figure}[htbp]
    \centering
    \includegraphics[width=1\textwidth]{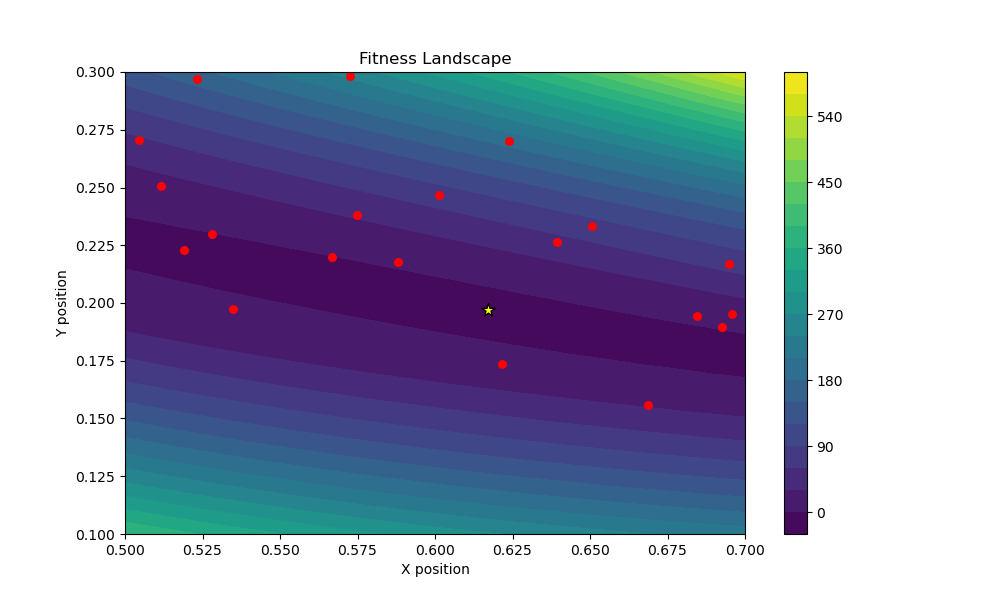}
    \caption{Illustration of the Negatively Correlated Search (NCS) framework. 
    The algorithm maintains $N$ parallel RLS processes, each of which can be regarded as a probability distribution $p_i(\boldsymbol{\theta})$.}
    \label{fig:NCS2}
\end{figure}

Within the ANTR framework, NCS is employed in each local trust region to generate diverse candidate solutions guided by the APT surrogate. Even when the surrogate posterior in a region tends toward unimodality, NCS ensures broad coverage of the parameter space by maintaining negatively correlated search behaviors. As shown in Fig.~\ref{fig:NCS2}, this design allows ANTR to obtain candidates that are not only competitive in fitness but also diverse in distribution, which helps the surrogate gradually correct its bias and mitigates the risk of premature convergence.

\subsection{Trust-Region Adaptation}\label{subsec3.4}

To address the heterogeneous structure and multimodality of the optimization space, this study adopts a trust region (TR) mechanism inspired by the TuRBO algorithm~\cite{eriksson2019scalable}. Unlike traditional Bayesian optimization, which typically relies on a single global surrogate and tends to converge to local optima in high-dimensional and multimodal settings, the trust region approach maintains multiple local regions to improve both exploration and adaptability.  

Specifically, in each iteration the algorithm maintains $M$ parallel trust regions $\{\mathcal{T}_1, \ldots, \mathcal{T}_M\}$, each represented as a hyper-rectangle in the parameter space and equipped with its own local surrogate model. At the first iteration, all trust regions share the global APT surrogate for evaluation. In subsequent iterations, each region independently trains its local surrogate using the historical data $\mathcal{D}$ restricted to samples falling within that region.  

To dynamically balance exploration and exploitation, the size of each trust region is adaptively adjusted according to its recent performance. If a region achieves consecutive improvements in three iterations (i.e., a newly evaluated solution outperforms the best one found so far), its size is doubled to encourage broader exploration. Conversely, if a region fails consecutively beyond a tolerance threshold $\text{fail}_{\text{tol}}$, its size is halved to emphasize local refinement. The threshold is defined as  
\begin{equation}
\text{fail}_{\text{tol}} = \left\lceil \max\left(\frac{4}{N}, \frac{d}{N} \right) \right\rceil,
\end{equation}
where $N$ is the population size and $d$ is the dimensionality of the parameter space.  

Through this adaptive mechanism, the optimization process focuses computational resources on promising areas while shrinking or discarding ineffective regions, thereby improving evaluation efficiency and accelerating convergence.

\section{Experiments and Analysis}\label{sec4}

\subsection{Objectives of the Experiments}\label{subsec4.1}

To evaluate the effectiveness of the proposed ANTR framework, we designed the experimental study with two primary objectives.
First, we conducted comparative experiments against representative evolutionary algorithms and SAEAs, with a particular focus on calibration accuracy measured by the mean squared error (MSE) and the parameter estimation error.
Second, we examined the internal compatibility of the proposed modules and assessed the robustness of ANTR when applied to datasets with varying characteristics, such as different model structures and time-series lengths.

\subsection{Experimental Setup}\label{subsec4.2}

This section introduces the experimental setup used for the calibration process and for evaluating the performance of the proposed ANTR framework.

\subsubsection{Simulation Models}\label{subsec4.2.1}

We considered two widely used nonlinear and agent-based models as simulation models:

\paragraph{Brock--Hommes Heterogeneous Expectations Model (BH model)}

The \textit{BH} model has been widely applied in financial market modeling and behavioral economics, as it captures how heterogeneous trading strategies and expectations jointly drive asset price dynamics. It is particularly effective in reproducing stylized financial phenomena such as price volatility, speculative behavior, and nonlinear market dynamics.

The model is formalized as the following coupled dynamical system:
\begin{equation}
x_{t+1} = \frac{1}{1 + r} \left[ \sum_{h=1}^H n_{h,t+1} (g_h x_t + b_h) + \epsilon_{t+1} \right],
\end{equation}
where $x_t$ denotes the asset price, $g_h$ and $b_h$ are the linear forecasting parameters of type-$h$ traders, $\epsilon_{t+1} \sim \mathcal{N}(0,\sigma^2)$ is a Gaussian noise term, and $r$ is the risk-free interest rate with $R=1+r$.

The fraction of type-$h$ traders at time $t+1$ is determined by a logit response function:
\begin{equation}
n_{h,t+1} = \frac{\exp\left(\beta U_{h,t}\right)}{\sum_{l=1}^H \exp\left(\beta U_{l,t}\right)},
\end{equation}
where $\beta$ is the intensity of choice parameter that governs traders’ sensitivity to utility differences, and $U_{h,t}$ represents the historical performance-based utility of type-$h$ traders, defined as
\begin{equation}
U_{h,t} = (x_t - R x_{t-1})(g_h x_{t-2} + b_h - R x_{t-1}).
\end{equation}

Through the nonlinear aggregation of multiple forecasting strategies, this model generates rich price dynamics. Even under simple parameter settings, it is capable of producing complex nonlinear behaviors such as cyclical fluctuations, speculative bubbles, and subsequent crashes.
In the experimental study presented in Section~\ref{subsec4.4.1}, the calibration tasks are defined as follows.  
When the problem dimension is $2$, the calibrated parameter vector is given by
\[
\boldsymbol{\theta} = [g_{2}, b_{2}], \quad g_{2}, b_{2} \in [0,1].
\]
When the dimension is $4$, the calibrated parameter vector becomes
\[
\boldsymbol{\theta} = [g_{2}, b_{2}, g_{3}, b_{3}], \quad g_{2}, b_{2}, g_{3} \in [0,1], \; b_{3} \in [-1,1].
\]

\paragraph{Preis-Golke-Paul-Schneider Model (PGPS)}

The \textit{PGPS} model is a representative agent-based model of limit-order-book (LOB) dynamics. It generates realistic market microstructure patterns driven by liquidity providers and takers, together with a stochastic mean-reversion mechanism governing directional order flow imbalance. The model captures market dynamics through the interaction of two types of agents, namely 125 liquidity providers and 125 liquidity takers. All agents share a common set of six hyperparameters to be calibrated, 
$
\boldsymbol{\omega} = [ \lambda_0, C_\lambda, \alpha, \mu, \Delta s, \delta],
$
whose meanings and corresponding parameter ranges are listed in Table~\ref{tab:pgps-para-range}.

At each time step \(t\), a liquidity provider submits a limit order with probability \(\alpha\), where the order volume is fixed to one unit. The probability that the order is placed on the bid or ask side is 0.5. Each liquidity taker submits a market order with probability \(\mu\), also with unit volume. A market order is placed on the bid or ask side with probabilities \(q_{\text{taker}}(t)\) and \(1-q_{\text{taker}}(t)\), respectively.

The variable \(q_{\text{taker}}(t)\) follows a mean-reverting random walk centered at 0.5. The probability of mean reversion is given by \(0.5 + |q_{\text{taker}}(t) - 0.5|\), and the step size toward the mean is \(\pm \Delta s\). In addition, liquidity takers cancel their unexecuted limit orders with probability \(\delta\).

Let \(p_a(t)\) and \(p_b(t)\) denote the best ask and bid prices at time \(t\). Market orders are executed at the best available price on the opposite side of the limit order book (LOB). The price of a limit order is determined by
$
p = p_s(t) + \lambda(t)\log(u) + s,
$
where \(u \sim U(0,1)\). For ask orders, \(p_s(t) = p_a(t)\) and \(s = 1\); for bid orders, \(p_s(t) = p_b(t)\) and \(s = -1\).

The time-varying order placement depth parameter \(\lambda(t)\) is defined as
\begin{equation}
\lambda(t) = \lambda_0
\left(
1 +
\frac{|q_{\text{taker}}(t) - \tfrac{1}{2}|}
{\langle q_{\text{taker}}(t) - \tfrac{1}{2} \rangle^2_{C_\lambda}}
\right),
\end{equation}
where \(\langle q_{\text{taker}}(t) - \tfrac{1}{2} \rangle^2\) is a precomputed normalization constant. Prior to the simulation, this value is estimated via \(10^5\) Monte Carlo iterations to ensure convergence to a reliable standard deviation.

\begin{table*}[htbp]
\small % 字体略缩小
\setlength{\tabcolsep}{4pt} % 默认是6pt，稍微缩小列间距
\caption{The ranges for randomly sampling parameters $\boldsymbol{\omega}$ to generate the synthetic data.}
\label{tab:pgps-para-range}
\begin{tabular*}{\textwidth}{@{\extracolsep\fill}ccc}
\toprule
\textbf{Parameter} & \textbf{Range} & \textbf{Remarks} \\
\midrule
$\lambda_{0}$   & [1, 200]      & Controls the base intensity of limit orders. \\
$C_{\lambda}$   & [1, 20]       & Scales the intensity of limit orders. \\
$\alpha$        & [0.05, 0.10]  & Probability that a liquidity provider submits a limit order. \\
$\mu$           & [0.10, 0.50]  & Probability that a liquidity taker submits a market order. \\
$\Delta s$    & [0.05, 0.10]  & Step size in the mean-reverting random walk of \(q_{\text{taker}}(t)\). \\
$\delta$        & [0.05, 0.10]  & Probability that a liquidity taker cancels an unexecuted order. \\
\botrule
\end{tabular*}
\end{table*}

\subsubsection{Training Data Generation and Embedding Network Architecture}\label{subsec4.2.2}

For training the global surrogate model on the Brock and Hommes heterogeneous expectations model, we constructed training datasets for parameter dimensions $d=2$ and $d=4$, where each parameter was defined over the range $[0,1]$. To ensure robustness in calibrating time series of varying lengths, we considered time series lengths ranging from 50 to 1000 with a step size of 10. For each time series length, Latin hypercube sampling was adopted to generate 200 parameter vectors within the parameter domain, resulting in a total of 19{,}200 training samples.

For the PGPS model, Latin hypercube sampling was similarly employed to generate a total of 20{,}000 parameter vectors. For each parameter vector, six time series with lengths $\{600, 1200, 1800, 2400, 3000, 3600\}$ were simulated, yielding a dataset comprising 120{,}000 parameter–time series pairs.

To handle heterogeneous sequence lengths, all simulated time series were padded to a unified maximum horizon $T_{\max}$ via zero-padding. During training, the padded regions were masked out in the loss computation and therefore did not affect gradient updates or parameter learning.

To embed the time series data, we employed a fully connected neural network as the feature extraction module. For the Brock and Hommes heterogeneous expectations model, a four-layer fully connected network was used to map padded time series of length 1000 into a 16-dimensional feature space. For the PGPS model, a deeper architecture consisting of five fully connected layers was adopted, mapping padded time series of length 3600 into a 64-dimensional feature space. ReLU activation functions were consistently applied throughout all network layers.

\subsubsection{Evaluation Metrics}\label{subsec4.2.3}

To quantitatively assess the performance of the calibration algorithms, we employed the following evaluation metrics:

\begin{enumerate}
    \item[1)] \textbf{Mean Squared Error (MSE):} This metric quantifies the discrepancy between the simulated time series $\hat{\boldsymbol{x}} = [\hat{x}_1, \hat{x}_2, \dots, \hat{x}_T]$ and the ground-truth data $\boldsymbol{x^*} = [x_1^*, x_2^*, \dots, x_T^*]$, and is defined as:
        \begin{equation}
        \text{MSE} = \frac{1}{T} \sum_{t=1}^{T} (x_t^* - \hat{x}_t)^2,
        \end{equation}
        where $T$ represents the length of the time series.
    
    \item[2)] \textbf{Parameter Estimation Error (Euclidean Distance):} This metric evaluates the accuracy of parameter estimation by measuring the Euclidean distance between the estimated parameter vector $\hat{\boldsymbol{\theta}} = [\hat{\theta}_1, \dots, \hat{\theta}_d]$ and the true parameter vector $\boldsymbol{\theta}^* = [\theta_1^*, \dots, \theta_d^*]$. It is computed as:
        \begin{equation}
        d(\hat{\boldsymbol{\theta}}, \boldsymbol{\theta}^*) = \sqrt{\sum_{i=1}^{d} (\hat{\theta}_i - \theta_i^*)^2},
        \end{equation}
    where $d$ represents the dimension of the parameter vector.

    \item[3)] \textbf{Calibration Success Rate:} Given a threshold $\varepsilon$,  a calibration run is considered successful if the Euclidean distance between the estimated and true parameter vectors satisfies the following criterion:
    \begin{equation}
    d(\hat{\boldsymbol{\theta}}, \boldsymbol{\theta}^*) \le \varepsilon = r \times \sqrt{d},
    \end{equation}
    where $d$ is the dimension of the parameter vector, and $r$ is a tolerance scaling factor. This criterion constrains acceptable solutions to lie within a hypersphere of radius $r\sqrt{d}$ in the normalized parameter space. The success rate is defined as the proportion of successful runs among all independent trials:
    \begin{equation}
    \text{Success Rate} = \frac{\text{Number of Successful Runs}}{\text{Total Number of Runs}} \times 100\%.
    \end{equation}
\end{enumerate}

\subsubsection{Computational Environment}\label{subsec4.2.3}

All experiments were conducted on a Linux server equipped with two Intel Xeon Platinum 8358 CPUs (2.60 GHz), providing 64 physical cores and 128 logical threads in total, and organized in a dual-socket NUMA architecture.

\subsection{Experimental Results}\label{subsec4.4}

\subsubsection{Experiments on the Brock and Hommes Heterogeneous Expectations Model}\label{subsec4.4.1}

\begin{figure}[htbp]
    \centering
    \includegraphics[width=0.85\textwidth]{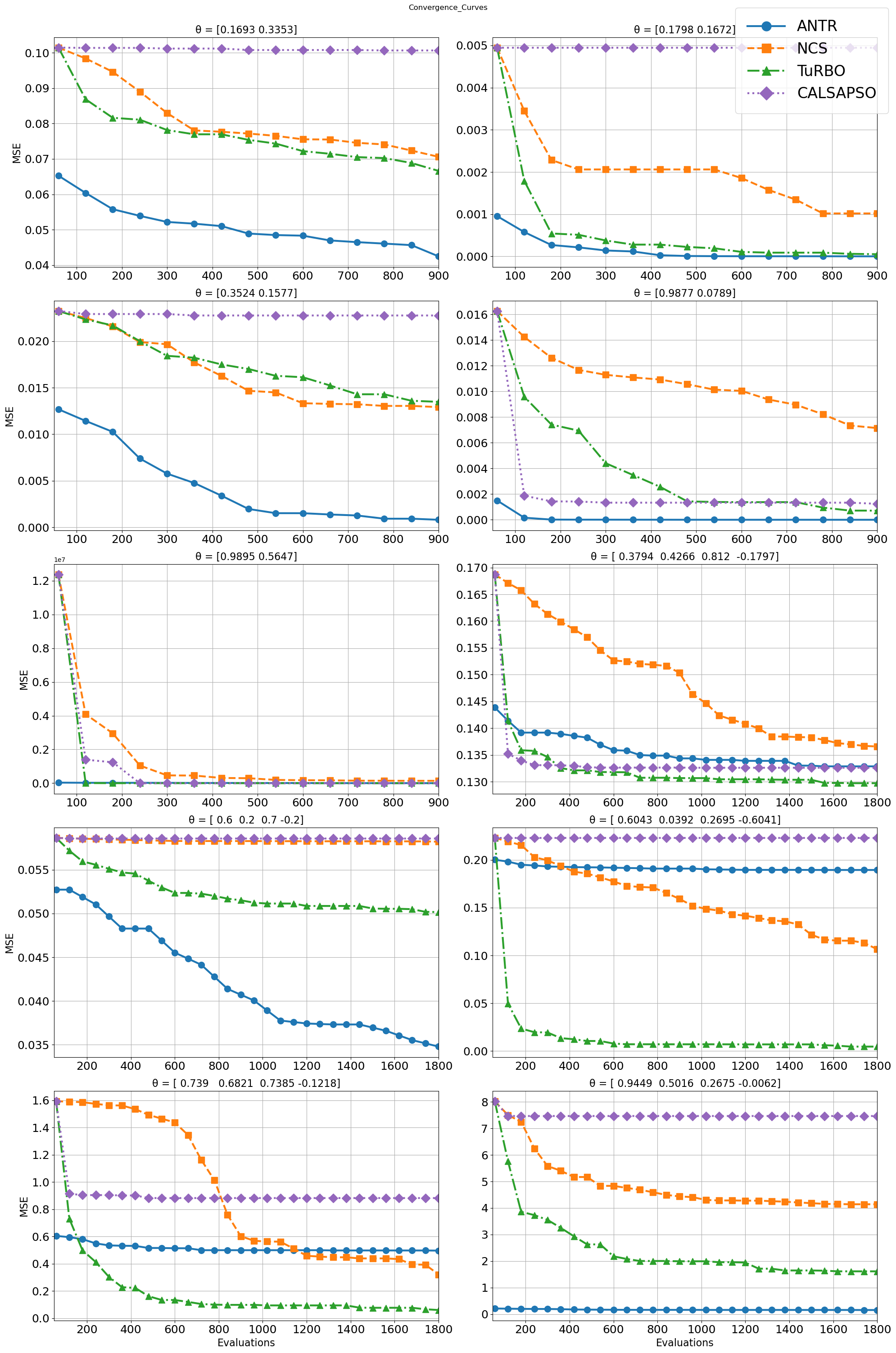}
    \caption{Comparison of MSE convergence curves on the Brock and Hommes Heterogeneous Expectations Model for ANTR, NCS, TuRBO, and CAL-SAPSO.}
    \label{fig:hommefunc-convergence}
\end{figure}

In this subsection, we evaluated the optimization performance and convergence behavior of the proposed ANTR method through comparative experiments. The baseline algorithms included NCS, TuRBO, and CAL-SAPSO. The experiments were conducted on the Brock and Hommes Heterogeneous Expectations Model, where the calibration targets consisted of 10 parameter vectors and the time series length was set to $T=900$. For problems with parameter dimension $d=2$, the maximum number of true evaluations was limited to 900, while for $d=4$ the limit was 1800. Each method was independently executed 10 times, and both the MSE and the Euclidean distance between the estimated and true parameter vectors were recorded. A calibration run was regarded as successful if the Euclidean distance was less than or equal to the threshold $\varepsilon = 0.1 \times \sqrt{d}$. The detailed results are reported in Tables~\ref{tab:hommefunc-mse}, \ref{tab:hommefunc-theta}, and \ref{tab:hommefunc-theta-success}, and are illustrated in Figure~\ref{fig:hommefunc-convergence}.

\begin{sidewaystable*}[htbp]
\centering
\caption{Comparison of MSE results on the Brock–Hommes  problems. Values are reported as mean $\pm$ standard deviation, and the best results are highlighted in bold.}
\label{tab:hommefunc-mse}
\begin{tabular*}{\textheight}{lllll}
\toprule
\textbf{Parameter} & \textbf{ANTR} & \textbf{NCS} & \textbf{TuRBO} & \textbf{CAL-SAPSO} \\
\midrule
(0.169, 0.335) & \textbf{4.25e-02 $\pm$ 8.73e-03} & 7.06e-02 $\pm$ 7.40e-03 & 6.64e-02 $\pm$ 7.59e-03 & 1.01e-01 $\pm$ 1.77e-03 \\
(0.180, 0.167) & \textbf{1.59e-06 $\pm$ 2.01e-06} & 1.02e-03 $\pm$ 9.33e-04 & 5.22e-05 $\pm$ 4.68e-05 & 4.95e-03 $\pm$ 8.67e-19 \\
(0.352, 0.158) & \textbf{8.09e-04 $\pm$ 1.85e-03} & 1.29e-02 $\pm$ 3.55e-03 & 1.35e-02 $\pm$ 5.11e-03 & 2.28e-02 $\pm$ 1.03e-03 \\
(0.988, 0.079) & \textbf{2.32e-07 $\pm$ 2.85e-07} & 7.14e-03 $\pm$ 9.18e-04 & 7.12e-04 $\pm$ 6.54e-04 & 1.24e-03 $\pm$ 1.40e-03 \\
(0.990, 0.565) & 1.56e+03 $\pm$ 1.80e+03 & 1.42e+05 $\pm$ 1.82e+05 & 4.27e+01 $\pm$ 5.08e+01 & \textbf{4.91e-01 $\pm$ 5.59e-02} \\
(0.379, 0.427, 0.812, -0.180) & 1.33e-01 $\pm$ 2.51e-03 & 1.37e-01 $\pm$ 2.42e-03 & \textbf{1.30e-01 $\pm$ 2.39e-03} & 1.33e-01 $\pm$ 4.51e-03 \\
(0.6, 0.2, 0.7, -0.2) & \textbf{3.48e-02 $\pm$ 9.20e-03} & 5.82e-02 $\pm$ 3.34e-05 & 5.02e-02 $\pm$ 1.33e-03 & 5.86e-02 $\pm$ 1.18e-04 \\
(0.604, 0.039, 0.270, -0.604) & 1.89e-01 $\pm$ 6.14e-04 & 1.06e-01 $\pm$ 2.97e-02 & \textbf{3.94e-03 $\pm$ 2.20e-03} & 2.23e-01 $\pm$ 0.00e+00 \\
(0.739, 0.682, 0.739, -0.122) & 4.97e-01 $\pm$ 5.42e-02 & 3.21e-01 $\pm$ 1.28e-01 & \textbf{5.73e-02 $\pm$ 3.47e-02} & 8.83e-01 $\pm$ 1.80e-01 \\
(0.945, 0.502, 0.268, -0.006) & \textbf{1.56e-01 $\pm$ 7.52e-03} & 4.14e+00 $\pm$ 4.25e-01 & 1.61e+00 $\pm$ 1.17e+00 & 7.47e+00 $\pm$ 1.28e+00 \\
\midrule
Friedman-test  & 1.8 & 3.0 & 1.8 & 3.4 \\
Wilcoxon-test & --  & 8-0-2 & 6-0-4 & 8-0-2 \\
\botrule
\end{tabular*}
\end{sidewaystable*}

\begin{sidewaystable*}[htbp]
\centering
\caption{Comparison of parameter estimation errors (mean $\pm$ standard deviation) on the Brock–Hommes problems. The best results are highlighted in bold.}
\label{tab:hommefunc-theta}
\begin{tabular*}{\textheight}{@{\extracolsep\fill}lllll}
\toprule
\textbf{Parameter} & \textbf{ANTR} & \textbf{NCS} & \textbf{TuRBO} & \textbf{CAL-SAPSO} \\
\midrule
(0.169, 0.335) & \textbf{1.26e-02 $\pm$ 2.21e-02} & 7.30e-02 $\pm$ 4.23e-02 & 1.20e-01 $\pm$ 4.71e-02 & 2.41e-01 $\pm$ 3.75e-02 \\
(0.180, 0.167) & \textbf{4.55e-04 $\pm$ 3.18e-04} & 1.33e-02 $\pm$ 1.01e-02 & 3.22e-03 $\pm$ 2.26e-03 & 2.63e-02 $\pm$ 0.00e+00 \\
(0.352, 0.158) & \textbf{5.41e-04 $\pm$ 3.89e-04} & 3.02e-02 $\pm$ 4.72e-02 & 2.71e-02 $\pm$ 2.09e-02 & 7.56e-02 $\pm$ 2.63e-02 \\
(0.988, 0.079) & \textbf{3.29e-04 $\pm$ 2.58e-04} & 1.45e-01 $\pm$ 5.73e-02 & 2.95e-02 $\pm$ 2.57e-02 & 2.09e-02 $\pm$ 1.56e-02 \\
(0.990, 0.565) & \textbf{4.85e-02 $\pm$ 5.00e-02} & 2.66e-01 $\pm$ 4.34e-02 & 2.67e-01 $\pm$ 1.20e-01 & 1.79e-01 $\pm$ 6.16e-03 \\
(0.379, 0.427, 0.812, -0.180) & \textbf{1.64e-01 $\pm$ 4.43e-02} & 5.72e-01 $\pm$ 1.12e-01 & 5.99e-01 $\pm$ 1.80e-01 & 8.07e-01 $\pm$ 0.00e+00 \\
(0.6, 0.2, 0.7, -0.2) & \textbf{9.05e-02 $\pm$ 1.22e-01} & 2.70e-01 $\pm$ 8.70e-03 & 3.80e-01 $\pm$ 4.81e-02 & 6.38e-01 $\pm$ 1.47e-01 \\
(0.604, 0.039, 0.270, -0.604) & 3.98e-01 $\pm$ 9.17e-02 & 4.65e-01 $\pm$ 1.08e-01 & \textbf{1.17e-01 $\pm$ 7.13e-02} & 3.99e-01 $\pm$ 5.55e-17 \\
(0.739, 0.682, 0.739, -0.122) & \textbf{9.94e-02 $\pm$ 2.92e-02} & 1.45e-01 $\pm$ 3.84e-02 & \textbf{8.61e-02 $\pm$ 3.94e-02} & 1.06e+00 $\pm$ 1.28e-01 \\
(0.945, 0.502, 0.268, -0.006) & \textbf{2.26e-01 $\pm$ 1.08e-01} & 5.58e-01 $\pm$ 2.19e-01 & 4.64e-01 $\pm$ 1.58e-01 & 7.29e-01 $\pm$ 1.41e-01 \\
\midrule
Friedman-test  & 1.2 & 2.9 & 2.5 & 3.4 \\
Wilcoxon-test & --  & 10-0-0 & 8-0-2 & 10-0-0 \\
\botrule
\end{tabular*}
\end{sidewaystable*}

\begin{table*}[htbp]
\caption{Success rates of parameter calibration on the Brock–Hommes problems. A run is considered successful if the Euclidean distance between estimated and true parameters is less than $\varepsilon = 0.1 \times \sqrt{d}$.}
\label{tab:hommefunc-theta-success}
\begin{tabular*}{\textwidth}{lcccc}
\toprule
\textbf{Parameter} & \textbf{ANTR} & \textbf{NCS} & \textbf{TuRBO} & \textbf{CAL-SAPSO} \\
\midrule
(0.169, 0.335) & \textbf{10} & 9 & 5 & 0 \\
(0.180, 0.167) & \textbf{10} & \textbf{10} & \textbf{10} & \textbf{10} \\
(0.352, 0.158) & \textbf{10} & 9 & \textbf{10} & \textbf{10} \\
(0.988, 0.079) & \textbf{10} & 3 & \textbf{10} & \textbf{10} \\
(0.990, 0.565) & \textbf{10} & 0 & 1 & 0 \\
(0.379, 0.427, 0.812, -0.180) & 9 & 0 & 0 & 0 \\
(0.6, 0.2, 0.7, -0.2) & 8 & 0 & 0 & 0 \\
(0.604, 0.039, 0.270, -0.604) & 0 & 0 & \textbf{6} & 0 \\
(0.739, 0.682, 0.739, -0.122) & \textbf{10} & 4 & \textbf{10} & 0 \\
(0.945, 0.502, 0.268, -0.006) & 4 & 0 & 0 & 0 \\
\botrule
\end{tabular*}
\end{table*}

\begin{table*}[htbp]
\caption{Evaluation budget percentage required by ANTR to match or surpass the best performance achieved by baseline algorithms on BH model (values in \%). \textgreater100\% indicates failure to reach the baseline optimum within the full budget.}
\label{tab:hommefunc-theta-success}
\centering
\begin{tabular*}{\textwidth}{l@{\extracolsep{\fill}}ccc}
\toprule
\textbf{Parameter} & \textbf{NCS} & \textbf{TuRBO} & \textbf{CAL-SAPSO} \\
\midrule
(0.169, 0.335)                  & 0     & 0     & 0     \\
(0.180, 0.167)                  & 0     & 40    & 0     \\
(0.352, 0.158)                  & 0     & 0     & 0     \\
(0.988, 0.079)                  & 0     & 6.7   & 60    \\
(0.990, 0.565)                  & 0     & \textgreater100  & 6.7   \\
(0.379, 0.427, 0.812, -0.180)   & 30    & \textgreater100  & \textgreater100  \\
(0.6, 0.2, 0.7, -0.2)           & 0     & 13.3  & 0     \\
(0.604, 0.039, 0.270, -0.604)   & \textgreater100  & \textgreater100  & 0     \\
(0.739, 0.682, 0.739, -0.122)   & \textgreater100  & \textgreater100  & 0     \\
(0.945, 0.502, 0.268, -0.006)   & 0     & 0     & 0     \\
\botrule
\end{tabular*}
\end{table*}

As shown in Table~\ref{tab:hommefunc-mse}, ANTR consistently achieved superior calibration accuracy on most parameter settings compared with the baseline algorithms. In particular, ANTR attained the lowest MSE in 6 out of 10 problems, while TuRBO achieved the best performance on 2 problems and CAL-SAPSO on 1 problem. When the parameter dimension increased to $d=4$, ANTR remained competitive and delivered stable results, whereas NCS and CAL-SAPSO often suffered from large variance or even convergence failures. The Wilcoxon signed-rank test further indicated that ANTR significantly outperformed NCS and CAL-SAPSO, with win--tie--loss counts of 8--0--2 and 8--0--2, respectively, while showing comparable performance to TuRBO (6--0--4). The Friedman test also highlighted the advantage of ANTR, with an average rank of 1.8, tied for the best among all algorithms.

Although the improvements in terms of MSE are not always statistically significant, as emphasized in Section~\ref{sec1}, researchers are more concerned with whether the estimated parameters are closer to the true values. As shown in Table~\ref{tab:hommefunc-theta}, the Wilcoxon test results demonstrated clearer improvements compared to those based on MSE, and the Friedman rank improved from 1.8 to 1.2. Moreover, Table~\ref{tab:hommefunc-theta-success} shows that ANTR achieved a 100\% calibration success rate in 6 out of 10 problems and outperformed all baselines in 9 out of 10 problems in terms of success rate.

Table~\ref{tab:hommefunc-theta-success} further quantifies the evaluation efficiency of ANTR by reporting the percentage of the evaluation budget required for ANTR to match or surpass the best performance achieved by the baseline algorithms on the Brock--Hommes problems. Overall, ANTR is able to reach the baseline optima using only a very small fraction of the total evaluation budget in most instances, and in many cases achieves performance comparable to or better than the baselines immediately (0\%). Taken together, these results highlight the strong sample efficiency of ANTR in achieving high-quality parameter calibration under strictly limited evaluation budgets.

Taken together, these findings indicate that ANTR provides more accurate parameter-level calibration and faster convergence than traditional SAEAs. Notably, conventional SAEAs cannot be pretrained across datasets of varying sequence lengths, which often leads to less effective initialization. In contrast, ANTR delivers high-quality initial estimates, thereby requiring fewer iterations to achieve comparable calibration accuracy. Moreover, ANTR naturally supports parallel computation, which further reduces the overall calibration time. These results on the Brock–Hommes problems confirm the effectiveness of ANTR. In the following, we extend our evaluation to the PGPS model, a more complex ABM with expensive simulation characteristics, in order to assess the scalability of ANTR in more realistic settings.

\subsubsection{Experiments on the PGPS Model}\label{subsec4.4.2}

While the experiments on the Brock and Hommes Heterogeneous Expectations Model demonstrated the ability of ANTR to handle nonlinear dynamics and heterogeneous expectations in a relatively low-dimensional setting, this model did not fully reflect the high computational cost inherent in ABM simulations. To further validate the scalability and practicality of ANTR, we next considered the PGPS model, which represents a computationally expensive ABM. The transition from the Brock and Hommes Heterogeneous Expectations Model to the PGPS model thus allowed us to progressively evaluate ANTR, moving from simplified nonlinear economic dynamics to a more realistic and costly agent-based financial market simulation.

In this study, we implemented our framework on the Multi-Agent Exchange Environment (MAXE)~\cite{maxe}, 
an agent-based market simulator developed at the University of Oxford and released under the MIT License. 

To evaluate the calibration performance, we randomly selected ten ground-truth parameter vectors, 
denoted as $\boldsymbol{\theta}_{1}$--$\boldsymbol{\theta}_{10}$, with their detailed values provided in Table~\ref{tab:pgps-para-sets}.

\begin{table*}[htbp]
\caption{Parameter settings of generating 10 synthetic datasets with the PGPS model (rounded to three decimal places).}
\label{tab:pgps-para-sets}
\begin{tabular*}{\textwidth}{ccccccc}
\toprule
\textbf{Parameters} & $\lambda_{0}$ & $C_{\lambda}$ & $\alpha$ & $\mu$ & $\Delta_{S}$ & $\delta$ \\
\midrule
$\boldsymbol{\theta}_{1}$  & 104.901 & 9.609  & 0.085 & 0.185 & 0.091 & 0.085 \\
$\boldsymbol{\theta}_{2}$  & 126.109 & 19.007 & 0.059 & 0.452 & 0.083 & 0.075 \\
$\boldsymbol{\theta}_{3}$  & 138.597 & 17.903 & 0.096 & 0.303 & 0.099 & 0.075 \\
$\boldsymbol{\theta}_{4}$  & 143.404 & 7.148  & 0.099 & 0.404 & 0.098 & 0.072 \\
$\boldsymbol{\theta}_{5}$  & 156.436 & 8.167  & 0.083 & 0.137 & 0.079 & 0.061 \\
$\boldsymbol{\theta}_{6}$  & 166.124 & 16.743 & 0.079 & 0.221 & 0.071 & 0.063 \\
$\boldsymbol{\theta}_{7}$  & 187.615 & 14.689 & 0.072 & 0.442 & 0.060 & 0.051 \\
$\boldsymbol{\theta}_{8}$  & 28.168  & 17.598 & 0.094 & 0.198 & 0.065 & 0.055 \\
$\boldsymbol{\theta}_{9}$  & 42.634  & 15.973 & 0.069 & 0.376 & 0.080 & 0.095 \\
$\boldsymbol{\theta}_{10}$ & 91.326  & 12.336 & 0.098 & 0.186 & 0.055 & 0.096 \\
\botrule
\end{tabular*}
\end{table*}

Similar to the experiments in Section~\ref{subsec4.4.1}, we evaluated the calibration performance of different algorithms on the PGPS model in terms of the MSE, the Euclidean distance between the calibrated and true parameters, and the number of successful calibrations. Due to the higher complexity of PGPS, all algorithms exhibited a certain deviation between the calibrated and true parameters. Therefore, we relaxed the success threshold to $\varepsilon = 0.3 \times \sqrt{d}$, where $d$ denotes the parameter dimension.

The results showed that ANTR was slightly inferior to TuRBO in terms of MSE on only one problem; however, on this task, the parameters calibrated by ANTR were actually closer to the ground truth than those obtained by TuRBO. Comparing the MSE and parameter-level results revealed that, although ANTR did not achieve an order-of-magnitude improvement in MSE over the baselines, its calibrated parameters were significantly closer to the true values. This demonstrates that ANTR provides much better parameter-level calibration than conventional SAEAs.

Table~\ref{tab:hommefunc-theta-success} further quantifies the evaluation efficiency of ANTR by reporting the percentage of the evaluation budget required for ANTR to match or surpass the best performance achieved by the baseline algorithms on the Brock--Hommes problems. Overall, ANTR was able to reach the baseline optima using only a very small fraction of the total evaluation budget in most instances, and in many cases achieved performance comparable to or better than the baselines immediately (0\%). Taken together, these results highlighted the strong sample efficiency of ANTR in achieving high-quality parameter calibration under strictly limited evaluation budgets.

Compared with the Brock–Hommes problems, ANTR maintained strong sample efficiency on the more complex PGPS model, as shown in Table~\ref{tab:pgps-success-rate}. Across most instances, it reached or exceeded the baseline optima using only a small fraction of the total evaluation budget.

\begin{sidewaystable*}[htbp]
\centering
\caption{Comparison of calibration performance on the PGPS model. Values are reported as mean $\pm$ standard deviation, and the best results are highlighted in bold.}

\label{tab:pgps-result1}
\begin{tabular*}{\textheight}{@{\extracolsep\fill}ccccc}
\toprule
\textbf{Parameters} & \textbf{ANTR} & \textbf{NCS} & \textbf{TuRBO} & \textbf{CAL-SAPSO} \\
\midrule
$\boldsymbol{\theta}_{1}$  & \textbf{3.97e+04 $\pm$ 1.39e+03} & 4.50e+04 $\pm$ 2.24e+03 & 4.19e+04 $\pm$ 1.48e+03 & 4.12e+04 $\pm$ 3.67e+03 \\
$\boldsymbol{\theta}_{2}$  & \textbf{4.38e+03 $\pm$ 6.80e+02} & 4.55e+03 $\pm$ 2.19e+02 & 4.49e+03 $\pm$ 1.35e+02 & 4.83e+03 $\pm$ 3.31e+02 \\
$\boldsymbol{\theta}_{3}$  & \textbf{5.04e+04 $\pm$ 5.75e+03} & 6.81e+04 $\pm$ 8.64e+03 & 6.25e+04 $\pm$ 1.50e+04 & 1.07e+05 $\pm$ 3.78e+04 \\
$\boldsymbol{\theta}_{4}$  & \textbf{3.47e+03 $\pm$ 1.67e+02} & 3.91e+03 $\pm$ 2.02e+02 & 3.85e+03 $\pm$ 1.35e+02 & 3.97e+03 $\pm$ 1.96e+02 \\
$\boldsymbol{\theta}_{5}$  & \textbf{1.58e+05 $\pm$ 2.99e+04} & 1.70e+05 $\pm$ 1.93e+04 & 1.60e+05 $\pm$ 1.49e+04 & 1.88e+05 $\pm$ 2.30e+04 \\
$\boldsymbol{\theta}_{6}$  & \textbf{3.68e+04 $\pm$ 2.84e+03} & 3.70e+04 $\pm$ 4.46e+02 & 3.71e+04 $\pm$ 9.38e+02 & 3.73e+04 $\pm$ 1.21e+03 \\
$\boldsymbol{\theta}_{7}$  & \textbf{2.47e+03 $\pm$ 2.10e+02} & 2.54e+03 $\pm$ 1.39e+02 & 2.56e+03 $\pm$ 2.47e+02 & 2.80e+03 $\pm$ 3.49e+02 \\
$\boldsymbol{\theta}_{8}$  & \textbf{1.24e+03 $\pm$ 6.55e+01} & 1.40e+03 $\pm$ 5.92e+01 & 1.30e+03 $\pm$ 7.09e+01 & 1.36e+03 $\pm$ 1.19e+01 \\
$\boldsymbol{\theta}_{9}$  & 9.87e+02 $\pm$ 9.82e+01 & 1.06e+03 $\pm$ 5.88e+01 & \textbf{9.76e+02 $\pm$ 1.00e+02} & 1.01e+03 $\pm$ 3.70e+01 \\
$\boldsymbol{\theta}_{10}$ & \textbf{5.66e+03 $\pm$ 1.98e+02} & 6.24e+03 $\pm$ 8.73e+01 & 5.90e+03 $\pm$ 2.41e+02 & 5.99e+03 $\pm$ 1.93e+02 \\
\midrule
Friedman-test  & 1.1 & 3.2 & 2.2 & 3.5 \\
Wilcoxon-test & --  & 7-3-0 & 7-3-0 & 7-3-0 \\
\botrule
\end{tabular*}
\end{sidewaystable*}

\begin{figure}[htbp]
    \centering
    \includegraphics[width=0.85\textwidth]{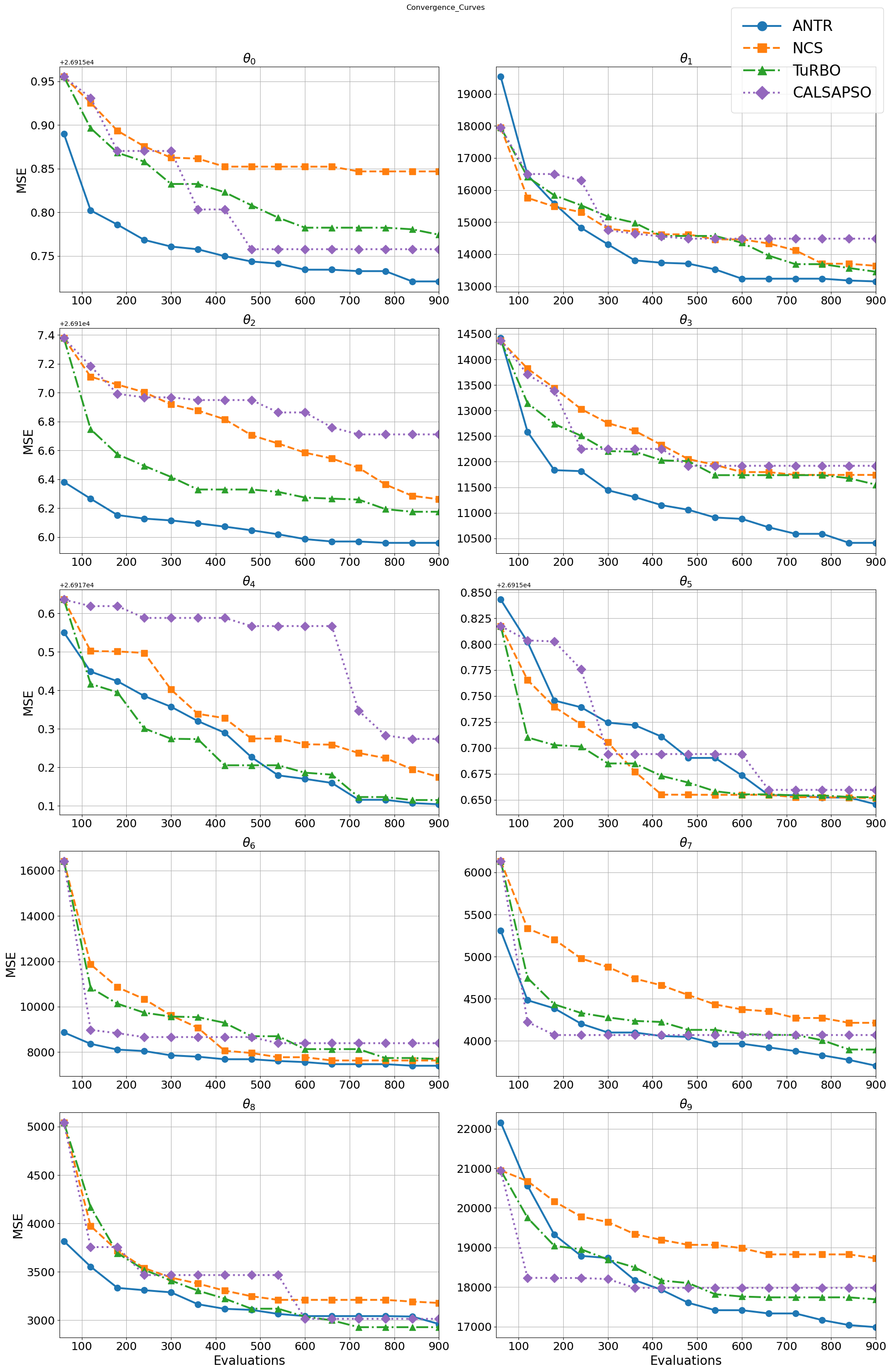}
    \caption{Comparison of MSE convergence curves on the MAXE model for ANTR, NCS, TuRBO, and CAL-SAPSO.}
    \label{fig:maxe-convergence}
\end{figure}

\begin{sidewaystable*}[htbp]
\centering
\caption{Comparison of parameter estimation errors (mean $\pm$ standard deviation) on the PGPS model. The best results are highlighted in bold.}
\label{tab:pgps-result2}
\begin{tabular*}{\textheight}{@{\extracolsep\fill}ccccc}
\toprule
\textbf{Parameters} & \textbf{ANTR} & \textbf{NCS} & \textbf{TuRBO} & \textbf{CAL-SAPSO} \\
\midrule
$\boldsymbol{\theta}_{1}$  & \textbf{5.62e-01 $\pm$ 7.27e-02} & 9.35e-01 $\pm$ 1.72e-01 & 7.16e-01 $\pm$ 1.82e-01 & 7.72e-01 $\pm$ 2.10e-01 \\
$\boldsymbol{\theta}_{2}$  & \textbf{4.14e-01 $\pm$ 1.36e-01} & 9.48e-01 $\pm$ 1.34e-01 & 1.13e+00 $\pm$ 1.39e-01 & 1.18e+00 $\pm$ 1.38e-01 \\
$\boldsymbol{\theta}_{3}$  & \textbf{4.50e-01 $\pm$ 7.71e-02} & 6.67e-01 $\pm$ 4.32e-01 & 1.00e+00 $\pm$ 1.14e-01 & 8.54e-01 $\pm$ 3.45e-01 \\
$\boldsymbol{\theta}_{4}$  & \textbf{3.58e-01 $\pm$ 5.08e-02} & 6.72e-01 $\pm$ 3.50e-01 & 7.88e-01 $\pm$ 1.06e-01 & 7.05e-01 $\pm$ 1.91e-01 \\
$\boldsymbol{\theta}_{5}$  & \textbf{3.26e-01 $\pm$ 1.20e-01} & 7.29e-01 $\pm$ 1.92e-01 & 6.39e-01 $\pm$ 1.42e-01 & 8.03e-01 $\pm$ 1.80e-01 \\
$\boldsymbol{\theta}_{6}$  & \textbf{2.99e-01 $\pm$ 6.64e-02} & 8.81e-01 $\pm$ 5.08e-02 & 8.13e-01 $\pm$ 8.46e-02 & 1.09e+00 $\pm$ 1.48e-01 \\
$\boldsymbol{\theta}_{7}$  & \textbf{6.87e-01 $\pm$ 1.10e-01} & 1.31e+00 $\pm$ 1.92e-01 & 1.08e+00 $\pm$ 1.56e-01 & 7.82e-01 $\pm$ 1.73e-01 \\
$\boldsymbol{\theta}_{8}$  & \textbf{3.31e-01 $\pm$ 6.84e-02} & 9.03e-01 $\pm$ 2.67e-01 & 8.84e-01 $\pm$ 3.13e-01 & 1.16e+00 $\pm$ 1.86e-01 \\
$\boldsymbol{\theta}_{9}$  & \textbf{6.43e-01 $\pm$ 1.21e-01} & 9.32e-01 $\pm$ 1.63e-01 & 1.04e+00 $\pm$ 1.59e-01 & 1.30e+00 $\pm$ 3.41e-02 \\
$\boldsymbol{\theta}_{10}$ & \textbf{7.48e-01 $\pm$ 1.18e-01} & 9.82e-01 $\pm$ 2.72e-01 & 9.09e-01 $\pm$ 1.79e-01 & 1.00e+00 $\pm$ 8.39e-02 \\
\midrule
Friedman-test  & 1.0 & 2.8 & 2.7 & 3.5 \\
Wilcoxon-test & --  & 9-1-0 & 9-1-0 & 9-1-0 \\
\botrule
\end{tabular*}
\end{sidewaystable*}

\begin{table}[htbp]
\caption{Success rates of parameter calibration on the PGPS model. A run is considered successful if the Euclidean distance between estimated and true parameters is less than $\varepsilon = 0.3 \times \sqrt{d}$.}
\label{tab:pgps-result3}
\begin{tabular*}{\textwidth}{@{\extracolsep\fill}ccccc}
\toprule
Parameters & ANTR & NCS & TuRBO & CAL-SAPSO \\
\midrule
$\boldsymbol{\theta}_{1}$  & \textbf{10} & 0 & 6 & 5 \\
$\boldsymbol{\theta}_{2}$  & \textbf{10} & 1 & 0 & 0 \\
$\boldsymbol{\theta}_{3}$  & \textbf{10} & 5 & 0 & 4 \\
$\boldsymbol{\theta}_{4}$  & \textbf{10} & 8 & 2 & 5 \\
$\boldsymbol{\theta}_{5}$  & \textbf{10} & 5 & 9 & 6 \\
$\boldsymbol{\theta}_{6}$  & \textbf{10} & 0 & 2 & 1 \\
$\boldsymbol{\theta}_{7}$  & 8 & 0 & 0 & 5 \\
$\boldsymbol{\theta}_{8}$  & \textbf{10} & 1 & \textbf{2} & 0 \\
$\boldsymbol{\theta}_{9}$  & 8 & 1 & \textbf{1} & 0 \\
$\boldsymbol{\theta}_{10}$ & 6 & 1 & 2 & 0 \\
\botrule
\end{tabular*}
\end{table}

\begin{table*}[htbp]
\caption{Evaluation budget percentage required by ANTR to match or surpass the best performance achieved by baseline algorithms on the PGPS model (values in \%). \textgreater100\% indicates failure to reach the baseline optimum within the full budget.}
\label{tab:pgps-success-rate}
\centering
\begin{tabular*}{\textwidth}{@{\extracolsep{\fill}}cccc}
\toprule
\textbf{Instance} & \textbf{NCS} & \textbf{TuRBO} & \textbf{CAL-SAPSO} \\
\midrule
$\boldsymbol{\theta}_{1}$  & 6.7   & 20    & 33.3 \\
$\boldsymbol{\theta}_{2}$  & 53.3  & 60    & 26.7 \\
$\boldsymbol{\theta}_{3}$  & 13.3  & 13.3  & 0    \\
$\boldsymbol{\theta}_{4}$  & 26.7  & 26.7  & 13.3 \\
$\boldsymbol{\theta}_{5}$  & 60    & 86.7  & 46.7 \\
$\boldsymbol{\theta}_{6}$  & 93.3  & 80    & 66.7 \\
$\boldsymbol{\theta}_{7}$  & 53.3  & 40    & 6.7  \\
$\boldsymbol{\theta}_{8}$  & 20    & 73.3  & 40   \\
$\boldsymbol{\theta}_{9}$  & 33.3  & \textgreater100   & 93.3 \\
$\boldsymbol{\theta}_{10}$  & 33.3  & 46.7  & 40   \\
\botrule
\end{tabular*}
\end{table*}

\subsubsection{Effectiveness of the NCS Module}\label{subsec4.4.3}

To assess the effectiveness of the NCS module, we conducted an ablation study designed to evaluate its contribution to enhancing exploration ability and preventing premature convergence to local optima. The Brock and Hommes heterogeneous expectations model was selected as the calibration benchmark. For ten distinct parameter configurations, we performed ten independent calibration runs each. The calibration results obtained by the ANTR algorithm (with NCS) and the ATR algorithm (without NCS) were recorded, and the Euclidean distance between the estimated parameter vector $\hat{\boldsymbol{\theta}}$ and the ground-truth parameter vector $\boldsymbol{\theta}^*$. The success threshold was set to $\varepsilon = 0.05 \times \sqrt{d}$.

As shown in Table~\ref{tab:ncs-effectiveness}, ANTR consistently achieved smaller Euclidean distances than ATR in the majority of test problems. Notably, ANTR attained a 100\% success rate (i.e., distance below $\varepsilon$) in 8 out of 10 cases, whereas ATR frequently failed to converge within the specified tolerance. To further examine the statistical significance of these differences, we applied the two-sided Wilcoxon signed-rank test and the Friedman test. Both tests rejected the null hypothesis at p < 0.05. The Friedman test indicated that ANTR (average rank = 1.1) performed significantly better than ATR (average rank = 1.9). Moreover, the Wilcoxon test showed that ANTR outperformed ATR in 8 problems, tied in 2, and was never inferior, thereby confirming the robustness of the improvements. These findings clearly demonstrate that the NCS module substantially enhances the exploration capability of the algorithm, effectively preventing premature convergence and improving calibration reliability. Consequently, incorporating NCS is essential for avoiding calibration results from being trapped in local optima.

\begin{sidewaystable*}[htbp]
\caption{Comparison between ANTR and ATR across 10 test problems. The best results are highlighted in bold.}
\label{tab:ncs-effectiveness}
\begin{tabular*}{\textheight}{@{\extracolsep\fill}ccccc}
\toprule
\textbf{Problem} & \textbf{ANTR} & \textbf{Success Rate} & \textbf{ATR} & \textbf{Success Rate} \\
\midrule
(0.034, 0.056) & 2.94e-01$\pm$1.95e-01 & 3  & \textbf{1.61e-01$\pm$1.60e-01} & 5  \\
(0.004, 0.060) & \textbf{1.30e-03$\pm$1.80e-03} & 10 & 2.02e-02$\pm$3.73e-03 & 10 \\
(0.105, 0.049) & \textbf{1.07e-01$\pm$1.52e-01} & 7  & 2.42e-01$\pm$7.31e-02 & 0  \\
(0.077, 0.048) & \textbf{2.57e-03$\pm$2.49e-03} & 10 & 9.28e-02$\pm$1.03e-01 & 6  \\
(0.043, 0.053) & \textbf{2.31e-03$\pm$1.43e-03} & 10 & 1.45e-01$\pm$1.58e-01 & 6  \\
(0.972, 0.001) & \textbf{2.65e-03$\pm$1.53e-03} & 10 & 9.18e-02$\pm$2.35e-01 & 9  \\
(0.129, 0.043) & \textbf{7.37e-04$\pm$5.29e-04} & 10 & 1.40e-02$\pm$4.52e-03 & 10 \\
(0.132, 0.045) & \textbf{1.33e-03$\pm$1.11e-03} & 10 & 8.22e-03$\pm$3.73e-03 & 10 \\
(0.124, 0.041) & \textbf{1.64e-03$\pm$1.36e-03} & 10 & 1.02e-02$\pm$3.57e-03 & 10 \\
(0.055, 0.049) & \textbf{1.51e-03$\pm$1.28e-03} & 10 & 3.87e-02$\pm$7.69e-02 & 9  \\
\midrule
Friedman-test  & 1.1 & -- & 1.9 & -- \\
Wilcoxon-test  & --  & -- & 8-2-0 & -- \\
\botrule
\end{tabular*}
\end{sidewaystable*}

\subsubsection{Compatibility on Different Sequence Lengths}\label{subsec4.4.4}

As discussed in Section~\ref{subsec3.2}, ANTR adopts zero-padding to extend all simulated output sequences to the maximum horizon $T_{\text{max}}$, thereby enabling batch calibration with heterogeneous sequence lengths. In this subsection, we investigate whether this preprocessing step introduces any bias in either the initialization or the final calibration results. To this end, we conducted experiments on $\boldsymbol{\theta} = [0.352, 0.158]$ using data lengths of $\{100, 300, 500, 700, 900\}$. As reported in Table~\ref{tab:robustness}, the Wilcoxon signed-rank test revealed no statistically significant differences across sequence lengths, for both the initial parameter estimates and the final calibrated results. These findings confirm that ANTR maintains robustness with respect to varying sequence lengths and that the zero-padding strategy does not compromise calibration accuracy.

\begin{table*}[htbp]
\small
\setlength{\tabcolsep}{4pt}
\caption{Robustness evaluation on different data lengths. Values are reported as mean $\pm$ standard deviation. 
The Wilcoxon-test results are reported in the form of win–tie–loss.}
\label{tab:robustness}
\begin{tabular*}{\textwidth}{@{\extracolsep\fill}ccc}
\toprule
\textbf{Length} & \textbf{Initial Estimate Parameters} & \textbf{Final Estimate Parameters} \\
\midrule
100 & [4.00e-01$\pm$8.20e-03, 1.41e-01$\pm$1.45e-02] & [3.52e-01$\pm$5.11e-04, 1.57e-01$\pm$3.44e-04] \\
300 & [4.08e-01$\pm$9.84e-03, 1.33e-01$\pm$1.68e-02] & [3.52e-01$\pm$5.63e-04, 1.58e-01$\pm$1.82e-04] \\
500 & [4.12e-01$\pm$1.39e-02, 1.38e-01$\pm$1.33e-02] & [3.51e-01$\pm$1.61e-03, 1.58e-01$\pm$2.67e-04] \\
700 & [4.16e-01$\pm$1.48e-02, 1.36e-01$\pm$1.20e-02] & [3.52e-01$\pm$5.14e-04, 1.58e-01$\pm$4.91e-04] \\
900 & [4.11e-01$\pm$1.36e-02, 1.42e-01$\pm$4.00e-03] & [3.52e-01$\pm$1.32e-03, 1.58e-01$\pm$1.63e-04] \\
\midrule
Wilcoxon-test & 0–5–0 & 0–5–0 \\
\botrule
\end{tabular*}
\end{table*}

\section{Conclusion}\label{sec5}

In this paper, we propose ANTR, a novel surrogate-assisted calibration framework for financial ABMs. By replacing conventional Gaussian process surrogates with a pretrained APT, ANTR directly models the posterior distribution over parameters, thereby resolving the mismatch between data-level fitness approximation and parameter-level estimation accuracy. To ensure robust exploration, we incorporate NCS and an adaptive trust-region mechanism into the framework. These components jointly enable effective multimodal search while concentrating computational resources on promising regions.

Extensive experiments on both the BH Model and the computationally expensive PGPS model demonstrate the effectiveness of the proposed approach. In particular, ANTR consistently achieves higher calibration accuracy than state-of-the-art SAEAs. The results further confirm that ANTR is robust to heterogeneous sequence lengths, scalable to higher-dimensional calibration tasks, and naturally supports parallel computation.

These findings underscore the potential of posterior-driven surrogate modeling to enhance the reliability of ABM calibration. By recovering parameters closer to their true values, ANTR increases the utility of calibrated models for downstream tasks such as policy evaluation, market intervention analysis, and risk assessment.

Nevertheless, several limitations remain. First, ANTR’s reliance on the trust-region mechanism to escape local optima offers room for improvement. Currently, the mechanism only triggers restarts when search stagnates and does not intelligently exploit information from previously evaluated samples to direct exploration toward more promising regions. Second, the NCS module does not yet exploit discrepancies between the surrogate landscape and the true objective landscape to adaptively select samples that are both informative and uncertain. Addressing these limitations could yield higher calibration accuracy and faster convergence, thereby enabling real-time online calibration.

Future work will focus on optimizing key components of the ANTR framework to achieve faster calibration and more reliable calibration performance.

\backmatter

\bmhead{Acknowledgements}

This work was supported by the National Natural Science Foundation of China (Grants 62272210, and 62331014).

\section*{Statements and Declarations}

\textbf{Funding}  
This work was supported by the National Natural Science Foundation of China (Grants 62272210, and 62331014).

\textbf{Competing Interests}  
The authors declare that they have no known competing financial or non-financial interests that could have appeared to influence the work reported in this paper.

\textbf{Data Availability}  
The data used in this study are available from the corresponding author upon reasonable request.

\textbf{Code Availability}  
The code used in this study is available from the corresponding author upon reasonable request.

\textbf{Author Contributions}  
All authors contributed to the study conception and design.

\bibliography{sn-bibliography}% common bib file

\end{document}